\documentclass[runningheads]{llncs}
\usepackage{graphicx}
\usepackage{tikz}
\usepackage{comment}
\usepackage{amsmath,amssymb} % define this before the line numbering.
\usepackage{color}

\usepackage{hyperref}
\hypersetup{
    colorlinks=true,            %链接颜色
    linkcolor=blue,             %内部链接
    filecolor=magenta,          %本地文档
    urlcolor=blue,              %网址链接
    pdftitle={Overleaf Example},
    pdfpagemode=FullScreen,
    }

\usepackage[accsupp]{axessibility}  % Improves PDF readability for those with disabilities.

\usepackage{amsmath}
\usepackage{algorithm}
\usepackage{algpseudocode}  
\usepackage{amsfonts}
\usepackage{enumerate}

\usepackage{ragged2e}
\usepackage{booktabs}
\usepackage{color}
\usepackage{multirow}
\usepackage{bm}
\usepackage{bbm}
\usepackage{graphicx}
\usepackage{subfigure}

\def\ie{{\em i.e.}}
\def\eg{{\em e.g.}}
\def\etal{{\em et al. }}

\newcommand{\figref}[1]{Fig. \ref{#1}}
\newcommand{\tabref}[1]{Tab. \ref{#1}}

\newcommand{\secref}[1]{Section \ref{#1}}
\newcommand{\myPara}[1]{\vspace{.05in}\noindent\textbf{#1}}

\newcommand{\mc}[1]{\mathcal{#1}}

\newcommand{\mb}[1]{\mathbb{#1}}

\newcommand{\tabincell}[2]{\begin{tabular}{@{}#1@{}}#2\end{tabular}}

\makeatletter
\def\thanks#1{\protected@xdef\@thanks{\@thanks
        \protect\footnotetext{#1}}}
\makeatother

\def\name{ReAct }
\def\att{RAID }
\def\cls{ACE }
\def\score{Segment Quality }

\begin{document}
% \renewcommand\thelinenumber{\color[rgb]{0.2,0.5,0.8}\normalfont\sffamily\scriptsize\arabic{linenumber}\color[rgb]{0,0,0}}
% \renewcommand\makeLineNumber {\hss\thelinenumber\ \hspace{6mm} \rlap{\hskip\textwidth\ \hspace{6.5mm}\thelinenumber}}
% \linenumbers
\pagestyle{headings}
\mainmatter
\def\ECCVSubNumber{4975}  % Insert your submission number here

%\title{ReAct: Temporal Action Detection with Relational Action Queries }
\title{ReAct: Temporal Action Detection  with Relational Queries}
% Replace with your title

% INITIAL SUBMISSION 
\begin{comment}
\titlerunning{ECCV-22 submission ID \ECCVSubNumber} 
\authorrunning{ECCV-22 submission ID \ECCVSubNumber} 
\author{Anonymous ECCV submission}
\institute{Paper ID \ECCVSubNumber}
\end{comment}
%******************

% CAMERA READY SUBMISSION
% \begin{comment}
%\titlerunning{ReAct: Temporal Action Detection with Relational Action Queries}
\titlerunning{ReAct: Temporal Action Detection  with Relational Queries}
% If the paper title is too long for the running head, you can set
% an abbreviated paper title here
%
\author{Dingfeng Shi\inst{1}$^{\ast}$\thanks{* This work is done during an internship at JD Explore Academy.
\\\dag~Corresponding authors: \email{mathqiong2012@gmail.com} and \email{jiali@buaa.edu.cn}.} \and
Yujie Zhong\inst{2} \and
Qiong Cao\inst{3\dag} \and \\ Jing Zhang\inst{4} \and Lin Ma\inst{2} \and Jia Li\inst{1\dag} \and Dacheng Tao\inst{3,4}}

\authorrunning{D. Shi et al.}
% First names are abbreviated in the running head.
% If there are more than two authors, 'et al.' is used.
%
\institute{State Key Laboratory of Virtual Reality Technology and Systems, School of Computer Science and Engineering, Beihang University \and
Meituan Inc \and JD Explore Academy \and The University of Sydney\\
% \email{\{shidingfeng,jiali\}@buaa.edu.cn ~~~jaszhong@hotmail.com\\ \{mathqiong2012,forest.linma,dacheng.tao\}@gmail.com jing.zhang1@sydney.edu.au}
}
% \end{comment}
%******************
\maketitle

\begin{abstract}
This work aims at advancing temporal action detection (TAD) using an encoder-decoder framework with action queries, similar to DETR, which has shown great success in object detection. However, the framework suffers from several problems if directly applied to TAD: 
the insufficient exploration of inter-query relation in the decoder, the inadequate classification training due to a limited number of training samples, and the unreliable classification scores at inference. 
To this end, we first propose a relational attention mechanism in the decoder, which guides the attention among queries based on their relations.
Moreover, we propose two losses to facilitate and stabilize the training of action classification.
Lastly, we propose to predict the localization quality of each action query at inference in order to distinguish high-quality queries.
The proposed method, named ReAct, achieves the state-of-the-art performance on THUMOS14, with much lower computational costs than previous methods. Besides, extensive ablation studies are conducted to verify the effectiveness of each proposed component. The code is available at \url{https://github.com/sssste/React}.

\end{abstract}

\section{Introduction}
Temporal action detection (TAD) has been actively studied because of the deep learning era. 
%Early methods focus on generating action proposals by localizing action boundaries, such as BSN~\cite{lin2018bsn} and BMN~\cite{lin2019bmn}. These methods usually require a second step to process and refine the large number of proposals. 
%Another line of 
Inspired by the advance of one-stage object detectors~\cite{lin2017focal,tian2019fcos,feng2021tood},
many recent works focus on one-stage action detectors~\cite{lin2021learning}, which show excellent performance while having a relatively simple structure. 
%In particular, the anchor-free methods directly output the final action predictions on the input video clips, without using hand-crafted anchors.
On the other hand, DETR~\cite{carion2020end}, which tackles object detection in a Transformer encoder-decoder framework, attracted considerable attention.
In this work, we propose a novel one-stage action detector \name that is based on such a learning paradigm. Inspired by DETR, \name models action instances as a set of learnable action queries. These action queries are fed into the decoder as inputs, and they iteratively attend to the output features of the encoder as well as update their predictions. The action classification and localization are then predicted by two simple feedforward neural nets.

\begin{figure}[t]
    \centering
    \setlength{\abovecaptionskip}{-0.2cm}
    \includegraphics[width=\linewidth]{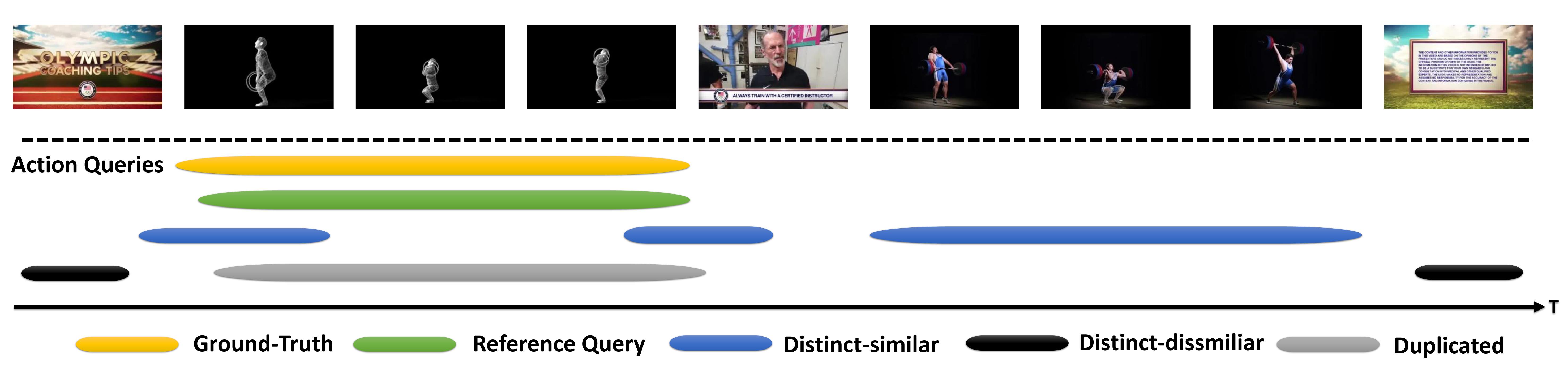}
    \label{intro}
  \caption{The relation of queries. We choose the green one as the reference query , and the queries in a different relation to it are labeled with different colors. Only the Distinct-similar pair (Blue ones) will be kept for attention computation.}
  \vspace{-0.5cm}
\end{figure}
% \vspace{-1.1cm}

However, the DETR-like methods suffer from several problems when applied to TAD task. 
First, the inter-query relations are not fully explored by the self-attention in the decoder, which is performed densely over all the queries.
Second, DETR-like methods may suffer from the inadequate training of action classification since the number of positive training samples for the classifier is relatively small compared to anchor-based/free methods. Moreover, when multiple queries fire for the same action instance at inference, queries with higher classification scores may not necessarily have better temporal localization.
In the following, we elaborate on these problems and introduce the proposed methods to alleviate them in three aspects: attention mechanism, training losses, and inference.

% 1st
The decoder in DETR-like methods applies the self-attention over the action queries to capture their relations, which can not fully explore the complex relations among queries.
In this work, we denote the action queries that are responsible for localizing different action instances of similar or same action classes as \emph{distinct-similar} queries, and those detecting different action classes as \emph{distinct-dissimilar} queries. For the queries that fire for the same action instance, we regard them as \emph{duplicate} queries.
In this work, we propose a novel \textbf{attention mechanism}, named Relational Attention with IoU Decay (RAID), to explicitly handle these three types of query relations in the decoder.
%For the distinct queries, we posit that similar queries (namely, ) provide more informative signals than dissimilar queries (\ie fire for different action classes). Therefore, the proposed relational attention focuses on the communication between similar queries and blocks the attention between dissimilar queries.
%On the other hand, duplicate queries with identical localization predictions do not contribute. Therefore, the proposed IoU decay encourages the duplicate queries to be slightly different by penalizing the IoU between queries to increase the probability of a more precise localization.
% 首页图
As \figref{intro} shows, \att focuses on the communication among distinct-similar queries (since they are expected to provide more informative signals) and blocks the attention between distinct-dissimilar and duplicate queries. Furthermore, the proposed IoU decay encourages the duplicate queries to be slightly different from each other to enable a more diverse prediction.

% 2nd
Another problem is that a DETR-like approach may have a relatively low classification accuracy due to inadequate classification training. This is because the positive training samples for the classification of DETR-like methods are much fewer than those of the anchor-free methods. Namely, 
%for anchor-free methods, all the temporal locations within the ground truth action segment~\cite{} (or locations near the center of the action~\cite{}) are considered as positives and contribute to the classification training. Whereas for 
for DETR-like methods, the number of positives per input clip is only the same as the ground truth actions because of the bipartite-matching-based label assignment.
To address this problem, we propose two \textbf{training losses}, codenamed Action Classification Enhancement (ACE) losses, to facilitate the classification learning. The first loss ACE-\emph{enc} is applied to the input features of the encoder and is designed to reduce the intra-class variance and inter-class similarity of action instances. This loss explicitly improves the discriminability of video features regarding acting classes, thus benefiting the classification. Meanwhile, a ACE-\emph{dec} loss is proposed as the classification loss in the decoder, which considers both the predicted segments and the ground-truth segments for action classification. It increases the training samples and generates a stable learning signal for the classifier.

% 3rd
Lastly, the action queries are redundant by design compared to the actual action instances. 
At inference, it is a common situation where multiple actions queries fire for the same action instance. 
Hence, it is important to focus on precise action localization queries. 
%For example, we need to prioritize the query with the most precise localization among its duplicate queries. 
Nonetheless, the classification score is deficient in measuring the temporal localization quality.
As a result, we propose a \score to predict the localization quality of each action query \textbf{at inference}, such that the more high-quality queries can be distinguished. 

To summarize, we make the following contributions in this work:

\begin{itemize}
    \item We approach temporal action detection using a DETR-like framework and identify three limitations of such method when directly applied to TAD.
    
    \item We propose the relational attention with IoU decay, the action classification enhancement losses, and the segment quality prediction, which alleviate the identified problems from the perspectives of attention mechanism, training losses, and network inference, respectively. 
    
    \item Experiments on two action detection benchmarks demonstrate the superiority of ReAct: it achieves the state-of-the-art performance on THUMOS14, with much lower computational costs than previous methods. Extensive ablation studies are conducted to verify the effectiveness of each component.
\end{itemize}

\section{Related Work}

% In order to assess our contribution in relation to the literature, it is necessary to consider three lines of research: temporal action detection, attention-based model, and contrastive learning.

\myPara{Temporal action detection.} Temporal action detection (TAD) aims to detect all the start and end timestamps and the corresponding action types based on the video stream information. The existing methods can be roughly divided into two categories: two-stage methods and one-stage methods. Two-stage methods~\cite{gao2018ctap,qing2021temporal,xu2020g,zeng2019graph,lin2019bmn,lin2018bsn,gao2017turn,lin2020fast} split the detection task into two subtasks: proposal generation and proposal classification. Concretely, some methods~\cite{lin2018bsn,lin2020fast,lin2019bmn} generate the proposals by predicting the probability of the start point and endpoint of the action and then selecting the proposal segments according to prediction score. In addition, PGCN~\cite{zeng2019graph} considers the relationship between proposals, then refines and classifies the proposals by Graph Convolutional Network. 
These two-stage methods can perform better by combining proposal generation networks and proposal classification networks. However, they can not be trained in an end-to-end manner and are computationally inefficient. To solve the above problems, some one-stage methods~\cite{lin2017single,chao2018rethinking,lin2021learning,long2019gaussian,xu2017r} are proposed.
% Lin~\etal ~\cite{lin2017single} builds detection networks with spatial and temporal features using 1D convolution.
Some works~\cite{chao2018rethinking,lin2021learning,yang2020revisiting} try to adapt to the high variance of the action duration by constructing a temporal feature pyramid, while Liu~\etal~\cite{liu2021end} propose to dynamically sample temporal features by learnable parameters. These one-stage methods reduce the complexity of the models, which are more computationally friendly. In this work, we mainly follow the one-stage fashion and the deformable convolution design~\cite{dai2017deformable,zhu2020deformable,liu2021end} to build a efficient action detector, which will be detailed in the \secref{sec:method}.

\myPara{Attention-based model.} 
Attention-based models~\cite{vaswani2017attention} have achieved great success in machine translation and been extended to the field of computer vision\cite{liu2021video,arnab2021vivit,liu2021swin,xu2021vitae,zhang2022vitaev2,chen2022dearkd} in recent years. The attention module computes a soft weight dynamically for a set of points at runtime. Concretely, DETR\cite{carion2020end} proposes a Transformer-based image detection paradigm. It learns decoder input features shared by all input videos and detects a fixed number of outputs. Deformable DETR~\cite{zhu2020deformable} improves DETR by reducing the number of pairs to be computed in the attention module with learnable spatial offsets. Liu~\etal~\cite{liu2021end} propose an end-to-end framework for TAD based on Deformable DETR. This type of training paradigm is highly efficient and fast in prediction. However, there is still a performance gap between these methods and the latest methods in TAD~\cite{liu2021end,zeng2019graph}. Our work is built on DETR-like workflows. In contrast to the above work, our approach suppresses the flow of invalid information by constricting a computational subset for the attention module, which improves performance effectively.

\myPara{Contrastive learning.}
Contrastive learning~\cite{chen2020simple} is a method that has been widely used in unsupervised learning. NCE~\cite{gutmann2010noise} mines data features by distinguishing between data and noise. Info-NCE~\cite{van2018representation} is proposed to extract representations from high-dimensional data with a probabilistic contrastive loss. Lin~\etal~\cite{lin2021learning} leverage contrastive learning to help network identify action boundaries. Inspired by these works, we use contrastive learning to extract a global common representation of action categories and enlarge the feature distance between action segments and noise segments. 

% 1. 提出了一个针对action detection的decoder：graph attention decoder。里面包括提出的graph attention和query IoU regularization

% 2. 提出了一个classification learning，来解决transformer-based模型在分类上表现不好的问题。这里包括了contrastive learning on features和gt-sample提供给分类器

% 3. 提出了query quality score来focus on high-quality queries。这里的quality score = iou * centerness

\section{Method}\label{sec:method}

\myPara{Problem definition.}
This work focuses on the problem of temporal action detection (TAD). Specifically, given a set of untrimmed videos $\mc{D}=\{\mc{V}_i\}_{i=1}^{n}$. A set of $\{X_i,Y_i\}$ can be extracted from each video $\mc{V}_i$, where $X_i=\{{x_t}\}_{t=1}^T$ corresponds to the image (and optical flow) features of T snippets and $Y_i=\{m_k,d_k,c_k\}_{k=1}^{K_i}$ is $K_i$ segment labels for the video $V_i$ with the action segment midpoint time $m_k$, the action duration $d_k$ and the corresponding action category $c_k$. Temporal action detection aims at predicting all segments $Y_i$ based on the input feature $X_i$. 

\begin{figure}[t]
    \centering
    \setlength{\abovecaptionskip}{-0.2cm}
    \includegraphics[width=\linewidth]{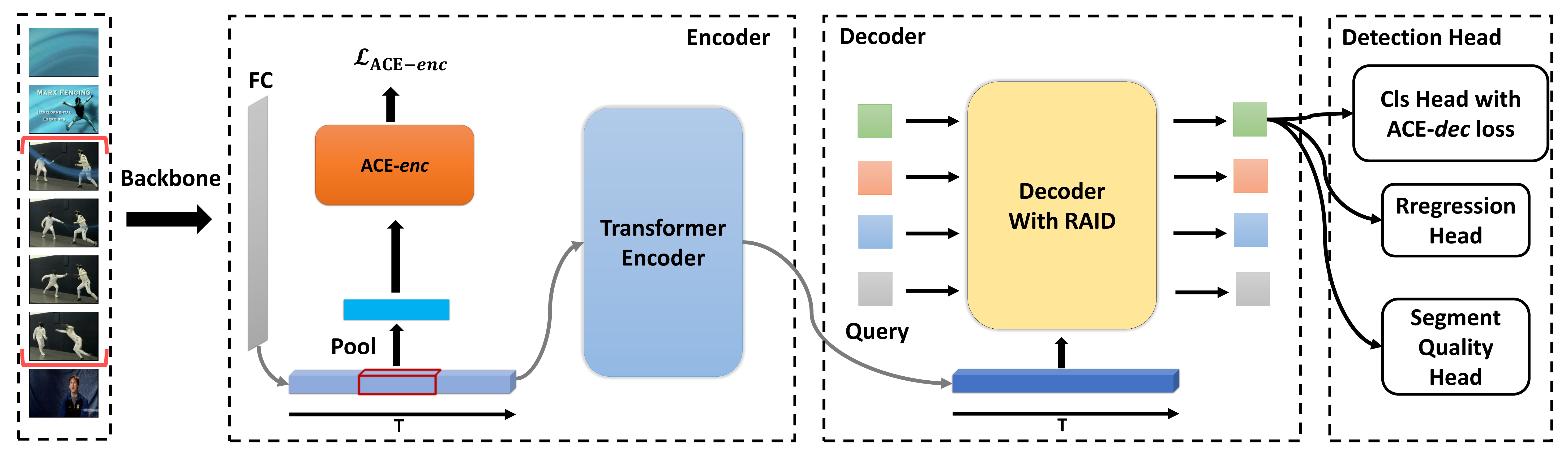}
    \label{framework}
    \caption{Illustration of the proposed framework. The video feature is extracted by a pretrained backbone, followed by a fully-connected layer to project the feature, and is additionally supervised by the AEC-Enc loss. After enhancement by the Transformer encoder, the features are fed into the decoder and attended by $L_q$ action queries in the decoder. The classification head is trained with the proposed ACE-Dec loss.} 
    \vspace{-0.4cm}
\end{figure}

\myPara{Method overview.}
Motivated by DETR~\cite{carion2020end}, we approach the problem of TAD by an encoder-decoder framework based on the transformer network. 
%Below, we first recap the preliminaries of DETR-like methods, and then introduce our proposed framework \name.
%\myPara{Preliminary.} 
As \figref{framework} shows, the overall architecture of \name contains three parts: a video feature extractor, an action encoder, and an action decoder.
First, video clip features are extracted from each RGB frame by using the widely-used 3D-CNN (\eg,  TSN~\cite{wang2018temporal} or I3D~\cite{carreira2017quo}). The optical flow features are also extracted using TVL1 optical flow algorithm~\cite{zach2007duality}.
Following that, a 1-D conv layer is used to modify the feature dimension of the clip features. The output features are then passed to the action encoder, which is a $L_E$-layer transformer network. The encoded clip features serve as one of the inputs to the action decoder. 
The decoder is a $L_D$-layer transformer, and it differs from the encoder in two aspects. It has action queries (which are learnable embeddings) as inputs, and the queries attend the encoder outputs in each layer of the decoder, known as Cross-attention.
Essentially, \name maps action instances as a set of action queries. The action queries are transformed by the decoder into output embeddings which are used for both action classification and temporal localization by separate feed-forward neural nets. 
The details of the encoder structure are provided in the appendix.

% \vspace{-0.4cm}
% \vspace{-1.1cm}

At training, following previous works\cite{carion2020end,zhu2020deformable,liu2021end}, the Hungarian Algorithm~\cite{kuhn1955hungarian} is applied to assign labels to the action queries. The edge weight is defined by the summation of the segment IoU, the probability of classification, and the L1 norm between two coordinates.
Based on the matching, \name applies several losses to the action queries, including the action classification loss and temporal segment regression loss.

%For the transformer decoder in deformable detr ~\cite{zhu2020deformable,liu2021end}, the input consists of $L_q$ query features from the encoder and its corresponding segments represented by the normalized midpoint and duration. Concretely, the query features are first fed into a Multi-Head Self-Attention (MHSA)~\cite{vaswani2017attention} module, followed by which a deformable cross-attention module predicts $K$ temporal sampling coordinates that is further normalized by referred segments. Then, $K$ features are sampled from the encoder output and summed up with an attention weight, which is also predicted by the input feature. Finally, fully connected layers are added on top and output features that will be used to generate new segments and action categories. 

%**preliminary only contains decoder description and is not clear enough，suggestion to use formulation or figure**

\myPara{Limitations of DETR-like methods for TAD.}
%DETR-like methods demonstrate good performance in object detection. However, they 
DETR-like methods may suffer from several problems when applied to TAD task. First, the decoder performs the self-attention densely over all the queries, which causes the inter-query relations not to be sufficiently explored. Second, compared with anchor-based/free methods, DETR-like methods may have issues in deficit training of action classification attributed to relatively smaller number of positive training samples for the classifier. Third, queries with higher classification scores may not be reliable due to multiple queries firing for the same action instance at inference. 

%In this work, we approach the problem of TAD by proposing an encoder-decoder based one-stage framework called \name. \figref{} demonstrates the overall architecture. Given the action instances represented by a set of learnable action queries and the output features of the encoder as input, \name allows these action instances relationally attending to the output features in conjunction with updating their prediction. Then, two simple feedforward neural nets are added for the final action classification and localization. 
In this work, we mitigate these problems in three aspects: (1) We propose the Relational Attention with IoU Decay which allows each action query to attend to others in the decoder based on their relations; (2) We design two Action Classification Enhancement losses to enhance the action classification learning at the encoder and decoder, respectively; (3) We introduce a \score to predict the localization quality of each action query at inference to compensate the deficiency of classification score at inference. We elaborate on these three aspects in the following. 

\begin{figure}[t]
    \centering
    \includegraphics[width=\linewidth]{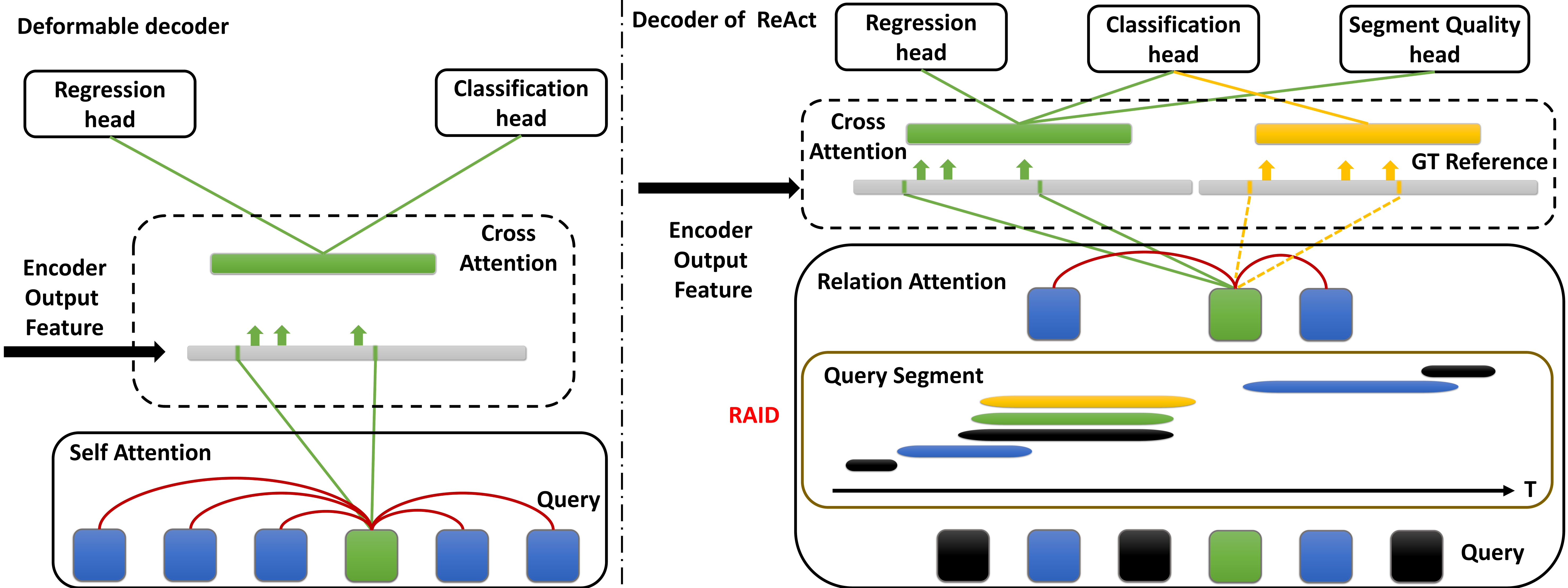}
    \caption{Illustration of our decoder. \textbf{Left:} plain deformable decoder. Each query performs the attention operation with all other query features and sample segment features from the encoder output. \textbf{Right:} decoder of ReAct. Each query only attends to specific queries based on the inter-query relation. Besides, the ground-truth segment provides an additional loss to further supervise the classification head. Note that for clarity, the LayerNorm, FFN, and residual connection are not shown in the figure (see appendix for detailed network structure).}
    \label{fig:decoder}
     \vspace{-0.4cm}
\end{figure}
%\vspace{-1.1cm}

\subsection{Relational Attention with IoU Decay}\label{RAID_sec} %{Multi-head Graph Attention Decoder}
To better explore the inter-query relation in the decoder, we present the Relational Attention with IoU Decay (RAID) which replaces the self-attention in the transformer decoder. 
%where the attention is effectively selected to a set of distinct queries which are responsible for localizing different action instances and the IoU decay encourages the duplicate queries that targets the same actions to slightly differ from each other.
Below, we describe the proposed method in detail.  

%**formulation need to be checked**

\myPara{Relational attention.} 
As a recap, we define three types of queries with respect to an action query $q_i$, which are differentiated by their relations to $q_i$.
\emph{Distinct-similar} queries are the queries that try to detect different action instances but of similar (or same) action class to $q_i$. 
\emph{Distinct-dissimilar} queries are those which try to detect different action instances and of dissimilar action class to $q_i$.
\emph{Duplicate} queries are the queries that try to detect the same action instance as $q_i$.
Intuitively, we anticipate that attending to distinct-dissimilar queries does not provide informative signals to $q_i$, since they focus on different action classes, and the relation between action classes may not be a reliable cue for detecting actions. On the contrary, attending to distinct-similar queries can benefit the query $q$ by gathering some background information and cues around $q_i$. For example, some actions may occur multiple times in a clip, and attending to each other can increase the confidence of the detection.
Moreover, duplicate queries only repeat the prediction as $q_i$, so they bring no extra information and should be ignored in the attention for $q_i$.

To find the distinct-similar queries for a query $q_i$, we consider two properties, namely, high context similarity and low temporal overlap.
To measure context similarity, we compute a similarity matrix $A\in\mb{R}^{L_q\times L_q}$ ($L_q$ is the number of queries) based on the query features, where each element represents the cosine similarity of two queries. Then the query-pair set $\mc{E}_{sim}$ is constructed by

\begin{equation}
\mc{E}_{sim}=\{(i,j) | A[i,j]-\gamma>0\},
% a^2 + b^2 = c^2 \label{num1}
\end{equation}
where $\gamma \in [-1,1]$ is a preset similarity threshold.
To identify the queries having low temporal overlap with $q$, a natural strategy is using the Interaction of Union (IoU) in the time domain, which measures the overlap between two temporal segments. Therefore, we compute a pair-wise IoU matrix $B \in\mb{R}^{L_q \times L_q}$ for the reference segments and construct a query-pair set $\mc{E}_{IoU}$ as follows:          
\begin{equation}
\mc{E}_{IoU}=\{(i,j) | B[i,j]-\tau<0\},
% a^2 + b^2 = c^2 \label{num1}
\end{equation}
where $i$ and $j$ denote the i-th and j-th queries respectively, and $\tau \in[0,1]$ is a preset IoU threshold. As shown in \figref{fig:decoder}, this simple strategy removes the segments which have large temporal overlap. 
%However, segments with small IoU may belong to similar or same class instances, and thus are not distinct with each other. To remedy this, we consider another source which provides similarity information based on the input features. 
We can then define the distinct-similar query-pair set $\mc{E}$ by combining $\mc{E}_{sim}$, $\mc{E}_{IoU}$ and the query itself $\mc{E}_s$. The definition is given as follow: 
\begin{equation}
\mc{E}= (\mc{E}_{IoU} \setminus \mc{E}_{sim} )\cup \mc{E}_s.
\end{equation}

% we add self-loops edge $\mc{E}_s=\{(i,i)\}_{i=1}^{L_q}$, 

% Similar to GAT\cite{velivckovic2017graph}, the node features are updated by weighted summation
% \begin{equation}
% F_i'=\sum_{j\in\mc{N}_i}{a_{ij}F_j},
% \end{equation}
% where $\mc{N}_i$ is the neighbors of node i, and the attention weight $a_{ij}$ is
% \begin{equation}
% a_{ij}=\frac{e^{F_i^TF_j}}{\sum_{k\in\mc{N}_i}{e^{F_i^TF_k}}}.
% \end{equation}

For a query $q_i$ and its distinct query-pair set $\mc{E}_i$, the key and value features can be written as $K_i=concatenate(\{k_j | (i,j) \in \mc{E}_i\})$ and $V_i=concatenate(\{v_j | (i,j) \in \mc{E}_i\})$. Then, the query features $q_i$ are updated by

\begin{equation}
    q_i' = a_iV_i^T,
\end{equation}
where the attention weight $a_i$ is
\begin{equation}
    a_i = Softmax_{K}(q_i K_i^T).
\end{equation}
Note that by considering both the context similarity and temporal overlap, the proposed relational attention successfully preserves the communication between $q_i$ and useful queries while blocking that between uninformative ones. 
%Figure XX visualises the effectiveness.

\myPara{IoU decay.} 
%We consider the action queries targeting the same action instance as \emph{duplicate} queries. 
Apart from relational attention, we introduce a further improvement by handling duplicate queries.
%To handle the issue that duplicate queries with identical localization prediction make no contribution, 
Namely, we propose a regularization, termed as IoU decay, which is added to the network optimization. It is given as   

\begin{equation}
    \omega_d=\frac{1}{2}\sum_{i=1}^{L_q}\sum_{j=1}^{L_q}{IoU(s_i,s_j)}.
\end{equation}
%**short description of how it is used?**
During the detector training, it penalizes the IoU between queries, such that duplicate queries can be diversified and different from each other, which can increase the probability of obtaining a more precise localization for the target action instance.

\subsection{Action Classification Enhancement} \label{cls_section}
To combat the issue of the inadequate learning of classification when applying the DETR-like methods to TAD, we propose two Action Classification Enhancement (ACE) losses to boost the classification performance. 
% We present the details below.

\textbf{ACE-\emph{enc} loss.} We aim to enhance the features with respect to action classification in the phase of encoder by enlarging the similarity of inter-class action instances and reducing the variance between intra-class action instances. 
We posit that explicitly increasing the discriminability of the features on the action detection dataset in an early stage can also benefit the final action classification.
Specifically, we optimize the input features of the encoder using contrastive loss.

The positive and negative action instance pairs are constructed as follows. For a given ground-truth action segment $s_{g}$ and its category $c_g$ in a video $v_i$, we choose its positive instances by sampling the action segments of the same category $c_g$ from either the same or different videos. As for its negative instances, we choose them from two different sources: (1) segments of action categories different from $c_g$ and (2) segments that are completely inside the ground-truth segment, but their IoU is less than a specific threshold $\xi$. 

For a given segment $s$, we denote $x\in \mb{R}^{T \times D'}$ and $\widetilde{x} \in \mb{R}^{T \times D}$ as the pre-trained video feature and feature further projected by a fully-connected layer $l$ (i.e. $\widetilde{x}=l(x)$), respectively. Then, the segment feature after temporal RoI pooling ~\cite{xu2020g} can be denoted as $f=RoI(\widetilde{x},s) \in \mb{R}^{D}$. With the above definitions, the loss $\mc{L}_{ACE-\emph{enc}}$ is given by   

\begin{equation}
    \mc{L}_{ACE-\emph{enc}}=-\log{\frac{\exp(f^T f_p)}{\sum_{j\in\mc{D}}{\exp(f^T f_j)}}},
\end{equation}
where $f_p$ is a positive segment of $f$ and $\mc{D}$ is the index of $k$ random negative instances as well as a positive instance.

%To achieve this, we add an extra single fully-connected layer after the output of backbone for each frame and use Info-NCE~\cite{van2018representation} to train it.

\textbf{ACE-\emph{dec} loss.} 
%Due to the fact that the deformable cross-attention module in the decoder update the query feature based on either the predicted output of the encoder or that of previous decoder layer. However, the prediction of segment in the early phase of training may be inaccurate，which misleads the classification learning and thus causes the inadequate learning. To handle this issue, we consider to enhance the action classification by providing more accurate training samples by feeding the classifier with features corresponding to the ground-truth action segments. 
Anchor-based/free methods treat all (or multiple) the temporal locations within the ground truth action segment as positives (\ie, belonging to an action class rather than backgrounds) for training the action classifiers, whereas DETR-like methods have much fewer positives due to the bipartite matching at label assignment. We,  therefore, propose the ACE-\emph{dec} loss to train the action classifiers.

As \figref{fig:decoder} (right) shows, in the training phase, %each input query feature first predicted a reference segment (the range indicated by the green slash) in which the cross attention module sample frames for computing weighted summation to get the segment feature (the green bar). The segment feature will be fed into the regression head and classification head for prediction. Some of these queries will be matched by ground-truth segments by Hungarian Algorithm. For these queries, we addtionally sample frames (the sample offsets are predicted by input query feature) from its corresponding ground-truth segment (\ie, the yellow segment). Finally, 
an additional positive training sample is fed to the action classifiers for each query segment (\ie, the green one) matched with a ground-truth action instance. The additional positive is obtained by feeding the ground-truth segment (\ie, the yellow one) as a normal query segment to the cross-attention layer.
The details of the cross-attention layer are described in the supplementary material. 
%and supervised. For multi-layer decoder, the ground-truth segment is fed to each layer and update the query feature, as input feature of the next layer.

%In the training phase, we adopted the design with twin decoders (see \figref{}). One of the decoders is a main deformable decoder with \att module. Given $L_q$ embedding features, $L_q$ detected segments will be predicted through multiple docoder layers, while each layer will update the query feature based on the intermediate predicted reference segment. Then, by bipartite matching, the p dectections (\ie~the number of ground-truth segments.) are assgined with the ground-truth segment and action label. 

%Another twin decoder share the same weight, classification head and input embedding with the main decoder. The difference is that, matching ground-truth segments will be fed into each decoder layer to guide the update of query features. After that, the same labels supervise the output on the same positions. 

Concretely, every decoder layer is attached a ACE-\emph{dec} loss which is given by  
\begin{equation}
    \mc{L}_{ACE-\emph{dec}}= \mc{L}_{foc}^q + \mathbbm{1}_{y\neq \emptyset}[\mc{L}_{foc}^{gt}],
\end{equation}
where $\mc{L}_{foc}^q$ and $\mc{L}_{foc}^{gt}$ is the sigmoid Focal Loss\cite{lin2017focal} for the query and ground-truth classification loss respectively. Note that, only the queries that are matched to ground-truth segments will contribute the ground-truth classification loss.

% With ACE-\emph{enc} and ACE-\emph{dec} modules, the overall loss for \cls is given by 
% \begin{equation}
%     \mc{L}_{ACE}= \mc{L}_{ACE-\emph{enc}} + \mc{L}_{ACE-\emph{dec}}.
% \end{equation}

\subsection{Segment Quality Prediction}
%To remedy the unreliable classification scores due to the redundant action queries by design, an obvious thought is to find a good metric for measuring the temporal localization quality. 
To remedy the problem that classification score is unreliable for selecting the best query among a set of duplicate queries, we propose a \score to predict the localization quality of each action query at inference for distinguishing high-quality queries. 
%To achieve that, we employ a feed-forward network that regresses the \score based on the predicted segment and its matched ground truth segment.     
The proposed segment quality prediction considers both the midpoint of the segment as well as its temporal coverage on the action instance.

Concretely, given a predicted segment $s_q$ and its query feature $f_q$, we define $(\zeta_{1}, \zeta_{2})=\phi(f_q)$, where $\phi$ is a single fully-connected layer and $\zeta_{1}, \zeta_{2} \in[0,1]$. Then, a final quality value $\zeta$ is defined by $\zeta = \zeta_{1} \cdot \zeta_{2}$. \score is supervised by a two-dimensional vector composing of the offset of the predicted midpoint and its ground truth for localizing the midpoint precisely, and the IoU between the predicted segment and its closest ground-truth segment for accurate temporal localization and coverage. The overall loss is given by       

% \score is supervised from two aspects: (1) the offset of the predicted midpoint and its ground truth for localizing the midpoint precisely; (2) the IoU between the predicted segment and its closest ground-truth segment for accurate temporal localization and coverage. The overall loss is given by

% \begin{equation}
%     \mc{L}_\zeta=\sum{\left| {\zeta_{1}-\exp({-\frac{|m_q-m_{gt}|}{l_{gt}} })} \right| +\left|{\zeta_{2}-IoU(s_q,s_{gt})}\right|},
% \end{equation}

\begin{equation}
    \mc{L}_\zeta=\sum{\left| {\phi(f_q)-(\exp({-\frac{1}{{l_{gt}}}{|m_q-m_{gt}|} }), IoU(s_q,s_{gt}))} \right|_1},
\end{equation}where $m_q$ is the midpoint of the predicted segment, and $m_{gt}$, $l_{gt}$ are the midpoint and length of the ground-truth segment, respectively. At inference, $\zeta$ is multiplied with the classification score of the segment.

% $\zeta_{1}$ and $\zeta_{2}$ $(\zeta_{1}, \zeta_{2}$) is predicted by a feed-forward network, respectively. 
%Our proposed \score is for providing the localization quality score of each predicted query, which, at inference, is multiplied to the classification score to yield a final confidence score.

%

\subsection{Training Losses}
%Based on the proposed losses in \cls and \score modules and 
At training, based on the label assignment by the Hungarian Algorithm, \name is trained by the total loss as follow: 

\begin{equation}
    \mc{L}=\mc{L}_{ACE-\emph{enc}} + \mc{L}_{ACE-\emph{dec}} + \mc{L}_\zeta + \mc{L}_{reg}.
\end{equation} 
%where $\lambda$ is hyper-parameters. 
%the label assignment by the Hungarian Algorithm,  
Here, $\mc{L}_{reg}$ is the commonly used regression loss for TAD which regresses the midpoint and the duration of the detected segments using the summation of L1 distance and the generalized IoU distance~\cite{rezatofighi2019generalized} for the matched pair. We define each objective as follows:
\begin{equation}
    \begin{split}
         &\mc{L}_{reg}=\frac{1}{N_{c_{gt}\neq \emptyset}}
        \sum_{j\in L_q}\mathbbm{1}_{c_{gt}^{(j)}\neq \emptyset}[\gamma_1\mc{L}_{L1}^{(j)} + \gamma_2\mc{L}_{gIoU}^{(j)}],\\
         &\mc{L}_{L1}^{(j)}= |m_{gt}^{(j)}-m^{(j)}|+|d_{gt}^{(j)}-d^{(j)}|,\\
         &\mc{L}_{gIoU}^{(j)} = 1-gIoU(s_{gt}^{(j)},s^{(j)}),\\
    \end{split}
\end{equation} 
where $s^{(j)}=(m^{(j)},d^{(j)})$ is j-th detected segment represented by midpoint and the duration. $c_{gt}^{(j)}$ is a set of the ground-truth segments that $s^{j}$ is matched and $N_{c_{gt}\neq \emptyset}$ is the number of segments in $c_{gt}^{(j)}$. $s_{gt}^{(j)}=(m_{gt}^{(j)},d_{gt}^{(j)})$ is the matched ground-truth segment of $s^{j}$ and $s_{gt}^{(j)} \in c_{gt}^{(j)}$. In addition, we follow the segment refinement fashion~\cite{zhu2020deformable,liu2021end} to predict detections in each decoder layer, each of which will be updated by summing with the upper layer segment and re-normalizing it. In this way, each layer provides auxiliary classification loss $\mc{L}_{cls}'$ and regression loss $\mc{L}_{reg}'$, which  further helps the network training.

% \subsubsection{Inference.}
% During testing, the classfication head output is activated by sigmoid. The category with the highest score will be chosen and the score will be update by multiplying \score. Then all the predictions will be process with Soft-NMS\cite{bodla2017soft} to remove the redundant and low quality segments.

\section{Experiment}
We conduct experiments on two challenging datasets: THUMOS14~\cite{THUMOS14} and ActivityNet-1.3~\cite{caba2015activitynet}.

%\subsection{Datasets and Settings}
%We conduct experiments on two challenging datasets: THUMOS14~\cite{THUMOS14} and ActivityNet-1.3~\cite{caba2015activitynet}. THUMOS14 consists of 20 categories of sports action. It contains 413 untrimmed videos for temporal action detection, 200 training, and 213 testings. ActivityNet-1.3 composes 10,024 training, 4,926 validation, and 5044 test videos from 200 classes of daily activities. We use the standard mean average precision (mAP) for performance evaluation with different IoU thresholds as the metric in our experiments. Following the common setting, we report thresholds [0.3: 0.7: 0.1] for THUMOS14 and \{0.5, 0.75, 0.95\} and average mAP under tIoU thresholds [0:5 : 0:05 : 0:95] for ActivityNet-1.3, respectively.

\subsection{Implementation Details} 
\subsubsection{Architecture details.}
For THUMOS14, we set $L_q=40$, $L_E=2$, $L_D=4$ for the number of queries, encoder layer and decoder layer, respectively. Each deformable attention module samples 4 temporal offsets for computing the attention. The hidden layer dimension of the feedforward network is set to 1024, and the other hidden feature dimension in the intermediate of the network is all set to 256. The pair-wise IoU threshold $\tau$ and feature similarity threshold $\gamma$ in \cls module are set to 0.2 and 0.2, respectively. 
%\subsubsection{ActivityNet-1.3.} 
For ActivityNet-1.3, we set $L_q=60$, $L_E=3$, $L_D=4$, $\tau=0.9$, $\gamma=-0.2$. We sample 4 temporal offsets for the deformable module. 
For more implementation details including feature extraction and training details, please refer to the supplementary material.

\subsubsection{Optimization parameters and inference.} We train the \name with AdamW optimizer with a batch size of 16. The learning rate is set to $2 \times 10^{-4}$ and $1 \times 10^{-4}$ for THUMOS14 and ActivityNet-1.3 respectively. \name is trained for 15 epochs on THUMOS14 and 35 epochs on ActivityNet-1.3. At inference, the classification head output is activated by sigmoid. Then all the predictions will be processed with Soft-NMS\cite{bodla2017soft} to remove the redundant and low-quality segments.

\begin{table}[t]
\centering{
\caption{Comparison with the state-of-the-art methods on THUMOS14 dataset. We report the mean Average Precision (mAP) in different thresholds and the floating-point operations (FLOPs, G).}
\label{table:thumos14}
\setlength{\tabcolsep}{1.2mm}
\renewcommand{\arraystretch}{1.1}
\begin{tabular}{c|c|c c c c c c | c}
\toprule
Type& Method & 0.3 & 0.4 & 0.5 & 0.6 & 0.7 & Avg. & FLOPs\\
\midrule
\multirow{9}*{\tabincell{c}{Two-stage}}
& BSN\cite{lin2018bsn}  & 53.5 & 45.0 & 36.9 & 28.4 & 20.0 & 36.8 & 3.4\\
& BMN\cite{lin2019bmn}  & 56.0 & 47.4 & 38.8 & 29.7 & 20.5 & 38.5 & 171.0\\
& G-TAD\cite{xu2020g}  & 54.5 & 47.6 & 40.3 & 30.8 & 23.4 & 39.3 & 639.8\\
& TAL\cite{chao2018rethinking}  & 53.2 & 48.5 & 42.8 & 33.8 & 20.8 & 39.8 &-\\
& TCANet\cite{qing2021temporal}  & 60.6 & 53.2 & 44.6 & 36.8 & 26.7 & 44.3 & -\\
& CSA+BMN\cite{sridhar2021class}  & 64.4 & 58.0 & 49.2 & 38.2 & 27.8 & 47.5 & - \\
& P-GCN\cite{zeng2019graph} & 63.6 & 57.8 & 49.1 & - & - & - & 4.4\\
& RTD-Net\cite{tan2021relaxed} & 68.3 & 62.3 & 51.9 & 38.8 & 23.7 & 49.0 & - \\
& VSGN\cite{zhao2021video} & 66.7 & 60.4 & 52.4 & 41.0 & 30.4 & 50.2 & -\\
& ContextLoc\cite{zhu2021enriching}  & 68.3 & 63.8 & 54.3 & 41.8 & 26.2 & 50.9 & 3.1\\
\midrule
\multirow{6}*{\tabincell{c}{One-stage}}
& SSAD\cite{lin2017single} & 43.0 & 35.0 & 24.6 & - & - & - & -\\
& SSN\cite{yu2019temporal} & 51.9 & 41.0 & 29.9 & - & - & - & -\\
& A2Net\cite{yang2020revisiting}  & 58.6 & 54.1 & 45.5 & 32.5 & 17.2 & 41.6 & 30.4\\
& AFSD\cite{lin2021learning}  & 67.3 & 62.4 & 55.5 & 43.7 & 31.1 & 52.0 & 5.1\\
& TadTr\cite{liu2021end}  & 62.4 & 57.4 & 49.2 & 37.8 & 26.3 & 46.6 & 0.75\\
\cline{2-9}
& \textbf{ReAct} & \textbf{69.2} & \textbf{65.0} & \textbf{57.1} & \textbf{47.8} & \textbf{35.6} & \textbf{55.0} & \textbf{0.68}\\
\bottomrule
\end{tabular}
}
% \vspace{-0.2cm}
\end{table}

\subsection{Main Results}
On THUMOS14 (see \tabref{table:thumos14}), our \name achieves superior performance and suppresses the state-of-the-art one-stage and two-stage methods in mAP at different thresholds. In particular, \name achieves 55.0\% in the average mAP, which outperforms TadTR by a large margin, namely about the 9.4\% absolute improvement. Besides, we compare the computational performance during testing. We adopt Floating-point operations per second (FLOPs) per clip following the previous works. \cite{liu2021end,zhu2021enriching}. We can see that our model has FLOPS of $0.68 G$, which is $0.06 G$ lower than TadTr and much lower than all the other methods. Note that the FLOPS we report in the table does not include the computation of video feature extraction with backbone. For methods like AFSD, which fine-tunes the backbone and does feature extraction during testing, we ignore the computation of feature extraction and only report the FLOPs afterward.

\begin{table}[t]
\centering{
\caption{Comparison with the state-of-the-art methods on ActivityNet-1.3 dataset.}
\label{table:activitynet}
\setlength{\tabcolsep}{1.3mm}
\renewcommand{\arraystretch}{1.1}
\begin{tabular}{c|c | c c c c | c }
\toprule
Type & Method & 0.5 & 0.75 & 0.95 & Avg. & FLOPs(G)\\
\midrule
\multirow{6}*{\tabincell{c}{Actioness}}
&BSN\cite{lin2018bsn}  & 46.5 & 30.0 & 8.0 & 28.2 & - \\
&SSN\cite{yu2019temporal}  & 43.2 & 28.7 & 5.6 & 28.3 & -\\
&BMN\cite{lin2019bmn}  & 50.1 & 34.8 & 8.3 & 33.9 & 45.6\\
&G-TAD\cite{xu2020g}   & 50.4 & 34.6 & 9.0 & 34.1 & 45.7\\
&BU-TAL\cite{zhao2020bottom} & 43.5 & 33.9 & \textbf{9.2} & 34.3 & -\\
&VSGN\cite{zhao2021video}  & \textbf{52.3} & \textbf{35.2} & 8.3 & \textbf{34.7} & -\\
\midrule
\multirow{4}*{\tabincell{c}{Anchor-based}}
&TAL\cite{chao2018rethinking}  & 38.2 & 18.3 & 1.3 & 20.2 & -\\
&PGCN\cite{zeng2019graph} & 48.3 & 33.2 & 3.3 & 31.1 & 5.0 \\
&TCANet\cite{qing2021temporal}  & 52.3 & \textbf{36.7} & \textbf{6.9} & \textbf{35.5}& -\\
&AFSD\cite{lin2021learning}  & \textbf{52.4} & 35.2 & 6.5 & 34.3& 15.3\\
\midrule
\multirow{3}*{\tabincell{c}{DETR-based}}
&RTD-Net\cite{tan2021relaxed} & 47.2 & 30.7 & \textbf{8.6} &  30.8& -\\
&TadTr\cite{liu2021end}  & 49.1 & 32.6 & 8.5 & 32.3 & \textbf{0.38}\\
&\textbf{ReAct}  & \textbf{49.6} & \textbf{33.0} & \textbf{8.6} & \textbf{32.6} & \textbf{0.38}\\
\bottomrule
\end{tabular}
}
% \vspace{-0.2cm}
\end{table}

On ActivityNet-1.3, our method achieves comparable results to the state-of-the-art (See \tabref{table:activitynet}). The \name outperforms the other DETR-based methods while enjoying a low computational cost (\eg, 0.38G). The Actioness and Anchor-based methods tend to have higher performance compared with the DETR-based methods. One possible reason is that the DETR-based methods take learnable query embedding as input,  which is video-agnostic and only keeps statistical information. For a dataset 
with a large variance in action time, a query feature has to take both long and short action into account (See appendix for more details) and is prone to conflicts. 
% To the end, we consider introducing video-relevant priory in future work.

\begin{table}[t]
\centering{
\caption{Ablation study on three main components.}
\label{table:ablation}
\setlength{\tabcolsep}{1.2mm}
\renewcommand{\arraystretch}{1.2}
\begin{tabular}{c| c | c | c | c c c c c c}
\toprule
Method & RAID & ACE & SQ & 0.3 & 0.4 & 0.5 & 0.6 & 0.7 & Avg. \\
\midrule
\multirow{5}*{\tabincell{l}{Our\\Base}}
 & & & & 66.6 & 59.2 & 49.7 & 38.0 & 25.0 & 47.7 \\
% \cline{2-10}
% & $\surd$ &  &  & 68.7 & 62.4 & 53.2 & 41.4 & 29.6 & 51.0 \\
  &  & $\surd$ & $\surd$ & 66.6 & 61.5 & 53.7 & 43.4 & 31.2 & 51.3 \\
%  \cline{2-10}
 & $\surd$ & & $\surd$ & 67.0 & 62.6 & 54.4 & 44.0 & 32.2 & 52.1 \\
% \cline{2-10}
 & $\surd$ & $\surd$ & & 69.1 & 63.3 & 54.2 & 43.5 & 31.0 & 52.2 \\
% \cline{2-10}
 & $\surd$ & $\surd$ & $\surd$ & \textbf{69.2} & \textbf{65.0} & \textbf{57.1} & \textbf{47.8} & \textbf{35.6} & \textbf{55.0} \\
\bottomrule
\end{tabular}
}
% \vspace{-0.2cm}
\end{table}

\subsection{Ablation Study}
In this section, we conduct the ablation studies on the THUMOS14 dataset.

\subsubsection{Main components.}
We demonstrate the effectiveness of three proposed components in ReAct: RAID, ACE, and Segment Quality. From \tabref{table:ablation} (row 2 and row 5), we can see that compared with the plain deformable decoder layer, our \att brings about a 3.7\% absolute improvement in the average mAP, proving the effectiveness of the module by introducing the relational attention based on the defined distinct-similar, distinct-dissimilar and duplicated queries. Besides, from rows 4 and 5 of the Table, we see our \cls improves the average mAP performance by 2.9\%, which shows its effectiveness by designing new losses to enhance classification learning. Finally, from rows 3 and 5, the proposed \score achieves 2.8\% improvements in average mAP, which effectively estimates the predicted segments' quality at inference. 

\begin{figure}[t]
    \centering{
    \setlength{\abovecaptionskip}{-0.005cm}
    \includegraphics[width=\linewidth]{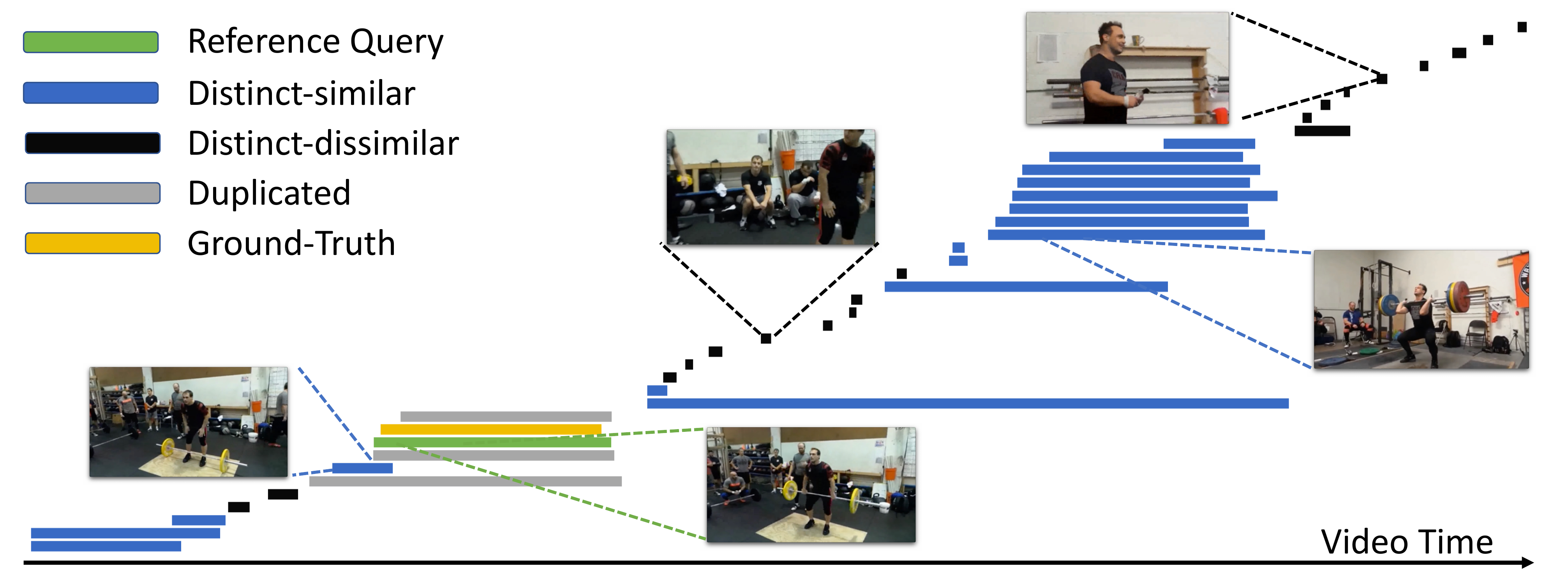}
    % \caption{Similarity}
    \label{fig:viz} 
    
  \caption{Visualization of the queries for a test video in THUMOS14. Some example frames are shown for the queries, and we can see that many distinct-dissimilar queries correspond to noises (i.e. not actions). }
  }
%   \vspace{-0.3cm}
\end{figure}

\subsubsection{Analysis of RAID.} 
\begin{figure}[t]
    \centering
    \setlength{\abovecaptionskip}{-0.005cm}
    \subfigure[Feature Similarity]{
    \includegraphics[width=0.46\linewidth]{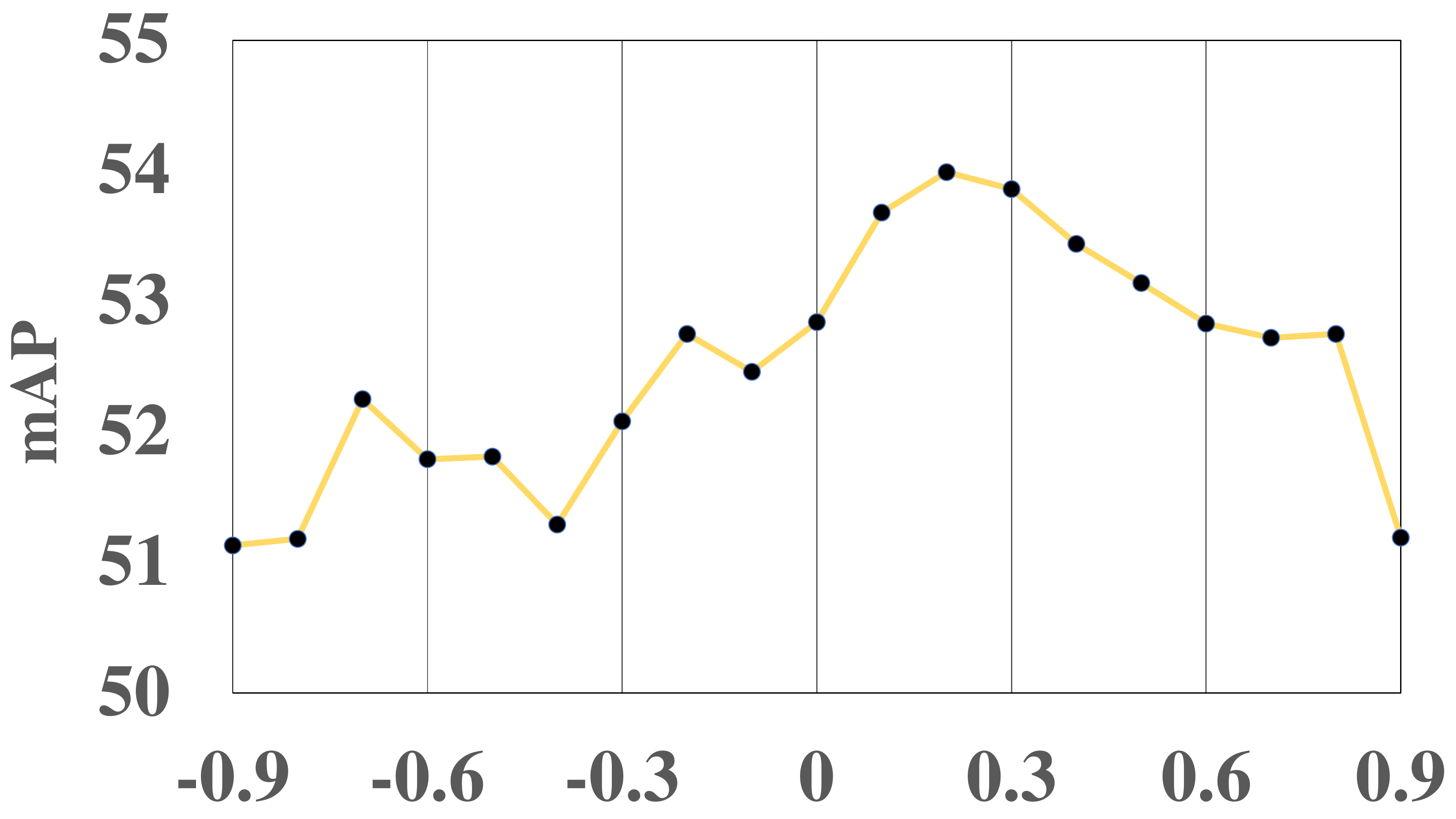}
    % \caption{Similarity}
    \label{fig:similarity} 
    }
    \quad
    \subfigure[IoU]{
    \includegraphics[width=0.46\linewidth]{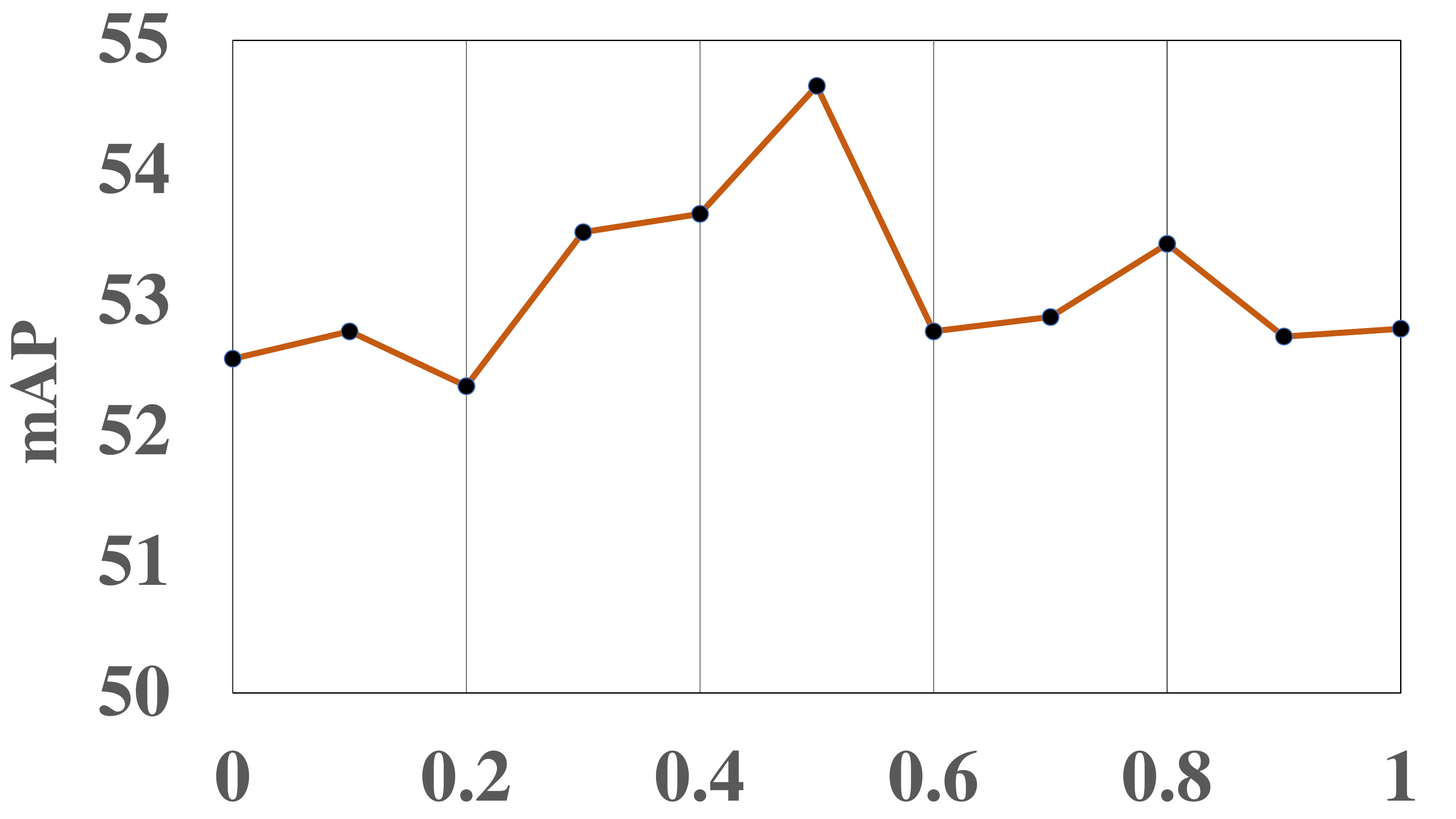}
    % \caption{IoU}
    \label{fig:iou} 
    }
   
  \caption{(a) is visualization of the choice of the hyperparameter $\gamma$, with $\tau = 1$; (b) is visualization of the choice of the hyperparameter $\tau$, with $\gamma = 0.2$ }
  \vspace{-0.3cm}
\end{figure}

We study the effect of two hyperparameters $\gamma$ and $\tau$ in \secref{RAID_sec} for thresholding the similarity scores and IoU values when constructing the distinct similar and dissimilar query sets. First, we set $\tau = 1$ and plot the average mAP when varying $\gamma$. From \figref{fig:similarity} we see that as $\gamma$ increases, the mAP exhibits an increase followed by a decrease, with a peak at $\tau=0.2$. Besides, we observe that the detection performance shows greater volatility as $\tau$ decreases further (\ie, $\tau<-0.1$). Intuitively, smaller $\tau$ leads to more irrelevant query pairs communicating, thus introducing greater uncertainty. Next, we study the effect of the choice of $\tau$ by fixing $\gamma = 0.2$. From \figref{fig:iou} we observe a similar trend with the figure for similarity as $\tau$ changes, and the optimal value is obtained at 0.5. Notice that the smaller $\tau$ is, the more queries will be excluded, and when $\tau=0$, only those that do not overlap will be retained. Intuitively, partially overlapped queries tend to be in the vicinity of the target query, which helps to perceive the information near the boundary. A visualized example of the queries is presented in \figref{fig:viz} to illustrate the work of RAID.

% the results are relatively stable as $\tau$ changes, 

\subsubsection{Analysis of ACE.}

\begin{table}[t]
\centering{
\caption{Comparison of different settings of ACE module.}
\label{table:cls}
\setlength{\tabcolsep}{1.2mm}
\renewcommand{\arraystretch}{1.2}
\begin{tabular}{c |c | c c c c c c}
\toprule
Module & Setting & 0.3 & 0.4 & 0.5 & 0.6 & 0.7 & Avg. \\
\midrule
\multirow{7}*{\tabincell{c}{ACE-\emph{enc}}}
& No Contrastive & 68.1 & 63.4 & 55.0 & 46.0 & 32.8 & 53.1 \\
\cline{2-8}
& \{S1,S2\} + \{N1\} & 68.3 & 63.4 & 55.4 & \textbf{46.2} & 33.9 & 53.4\\ 
& \{S1,S2\} + \{N2\} & 69.7 & 64.6 & 55.7 & 45.6 & 33.8 & 53.9\\
& \{S1,S2\} + \{N1,N2\} & \textbf{69.7} & \textbf{64.5} & \textbf{56.6} & 45.9 & \textbf{34.7} & \textbf{54.3}\\ 
& \{S1\} + \{N1,N2\} & 69.1 & 64.4 & 56.3 & \textbf{46.2} & 34.6 & 54.1 \\
\cline{2-8}
& Before Transformer Enc. & \textbf{69.7} & \textbf{64.3} & \textbf{56.1} & \textbf{46.4} & \textbf{34.2} & \textbf{54.1}\\ 
& After Transformer Enc. & 66.4 & 61.2 & 53.3 & 43.4 & 32.0 & 51.2 \\
\midrule
\multirow{3}*{\tabincell{c}{ACE-\emph{dec}}}
& $\mc{L}_{foc}^q$ Only  & 67.5 & 62.6 & 53.9 & 43.3 & 33.2 & 52.1 \\
& $\mc{L}_{foc}^{gt}$ Only & 66.1 & 61.1 & 53.6 & 44.2 & 30.9 & 51.2\\
& $\mc{L}_{foc}^q$ + $\mc{L}_{foc}^{gt}$ & \textbf{68.3} & \textbf{63.4} & \textbf{55.4} & \textbf{46.2} & \textbf{33.9} & \textbf{53.4}\\ 
\bottomrule
\end{tabular}
}
% \vspace{-0.5cm}
\end{table}

We analyze the effect of ACE-enc loss in the following aspects: the construction of contrastive pairs, where to apply ACE-\emph{enc} loss and training losses. First, we study how contrastive pairs affect performance. In particular, to form the positive segment pairs, we randomly choose segments of the same category from either the same video or different videos, denoted by \textbf{S1} and \textbf{S2}, respectively. As for negative pairs, there are two ways: segment pairs belonging to different action classes (denoted by \textbf{N1}), and segment pairs that one completely includes the other, but their IoU is less than a threshold (denoted by \textbf{N2}), as described in \ref{cls_section}. \tabref{table:cls} presents the results using different combinations of positive and negative pairs. In \tabref{table:cls}, we see that \textbf{N2} play a more important role in training than \textbf{N1} (\eg, average mAP 53.9 versus 53.4), and merging them can gain further promotion (\ie, 54.3). 

Secondly, we study the effect of where to apply ACE-\emph{enc} loss. We mainly consider two positions: before the transformer encoder and after it. We train a single fully connected layer for the former to enhance the video features. For the latter, we use the encoder output. The experimental results show that a single fully connected layer is much better than a complex transformer encoder. Intuitively, after encoder processing, the features on each frame already contain local temporal information, therefore, the pooled segment features can not represent the action precisely, leading to inaccurate convergence.

Finally, to go deeper into the ACE-\emph{dec} loss, we conducted three experiments: query classification loss only, ground-truth classification loss only, and the complete ACE-\emph{dec} loss. For the case of the ground-truth classification loss only, we still predict and match the ground-truth segment with the input query feature, which provides the matched query position and reference ground-truth segment. However, we only update the network with ground-truth classification loss $\mc{L}_{foc}^{gt}$. From the \tabref{table:cls}, neither $\mc{L}_{foc}^{q}$ nor $\mc{L}_{foc}^{gt}$ can perform well, but when we combine them together, the result are significantly better (\eg, 53.4 versus 51.2).

\section{Conclusion}
In this work, we consider the task of temporal action detection and propose a novel one-stage action detector \name based on a DETR-like learning framework. Three limitations of such a method when directly applied to TAD are identified. We propose the relational attention with IoU decay, the action classification enhancement losses, and the segment quality prediction and handle those issues from three aspects: attention mechanism, training losses, and network inference, respectively. \name achieves the state-of-the-art performance with much lower computational costs than previous methods on THUMOS14. Extensive ablation studies are also conducted to demonstrate the effectiveness of each proposed component. In the future, we plan to include the video feature extractor in the action detection training to improve the performance further. 

\subsubsection{Acknowledgement.} 
This work is supported by the Major Science and Technology Innovation 2030 "New Generation Artificial Intelligence" key project (No. 2021ZD0111700), National Natural Science Foundation of China under Grant 62132002, Grant 61922006 and Grant 62102206.

\clearpage

\appendix
% \renewcommand{\thefigure}{\Alph{figure}}
% \counterwithout{figure}{section}

\title{Supplementary Material For ReAct} % Replace with your title

\titlerunning{Supplementary Material For ReAct}
\author{}
\institute{}
\maketitle

\section{Encoder in Detail}
\begin{figure}[]
    \centering
    \setlength{\abovecaptionskip}{-0.2cm}
    \includegraphics[width=\linewidth]{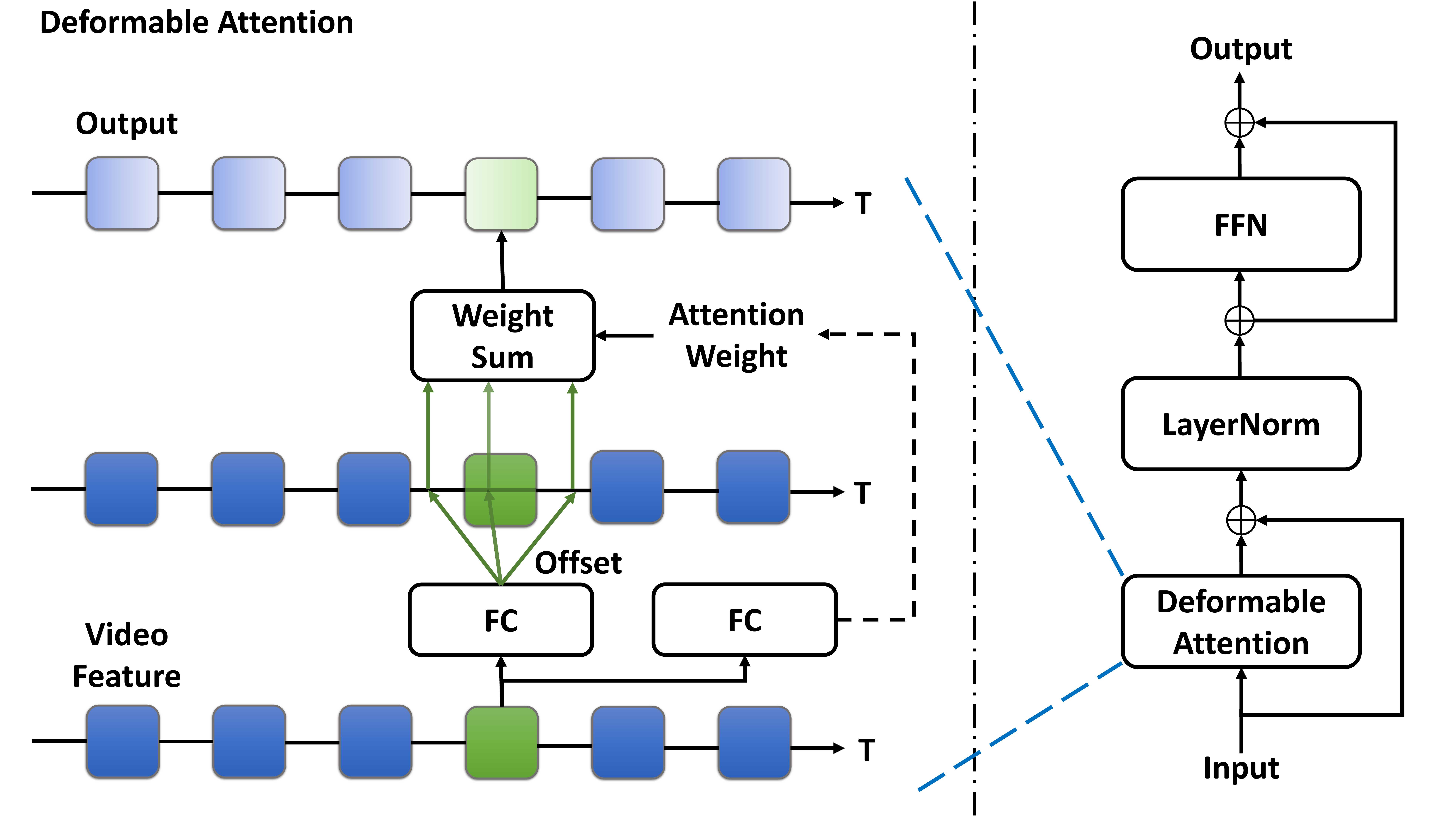}
  \caption{Illustration of the encoder.}
  \label{encoder}
  \vspace{-0.5cm}
\end{figure}

To be self-contained, we provide the detailed structure of the encoder.
As \figref{encoder} shows, for the input video feature $F\in\mb{R}^{T \times D}$, a local offset position and attention weight will be predicted with two fully-connected layers, respectively. For each time step, feature are then sampled according to the $K$ offsets with linear interpolation. The sampled features are weighted by the attention weights and summed up to produce the updated frame feature for the corresponding time step.

\vspace{-0.4 cm}
% \vspace{-1.1cm}

% \myPara{Position Embedding.} Following previous works~\cite{vaswani2017attention,liu2021end}, the input feature is added with position embedding given by
% $$Pos(t,d)=
% \begin{cases}
% \sin{\frac{t}{10000^{d/D}}}& \text{d is even}\\
% \sin{\frac{t}{10000^{(d-1)/D}}}& \text{d is odd},
% \end{cases}$$
% where $t$ and $d$ is the temporal dimension and feature dimension respectively.

\section{Decoder in Detail}
To help understand our method better, we introduce the decoder in detail. There are two attention modules in the decoder: the proposed relational attention module and a cross-attention module.

% \vspace{-0.4 cm}
\begin{figure}[]
    \centering
    \setlength{\abovecaptionskip}{-0.2cm}
    \includegraphics[width=\linewidth]{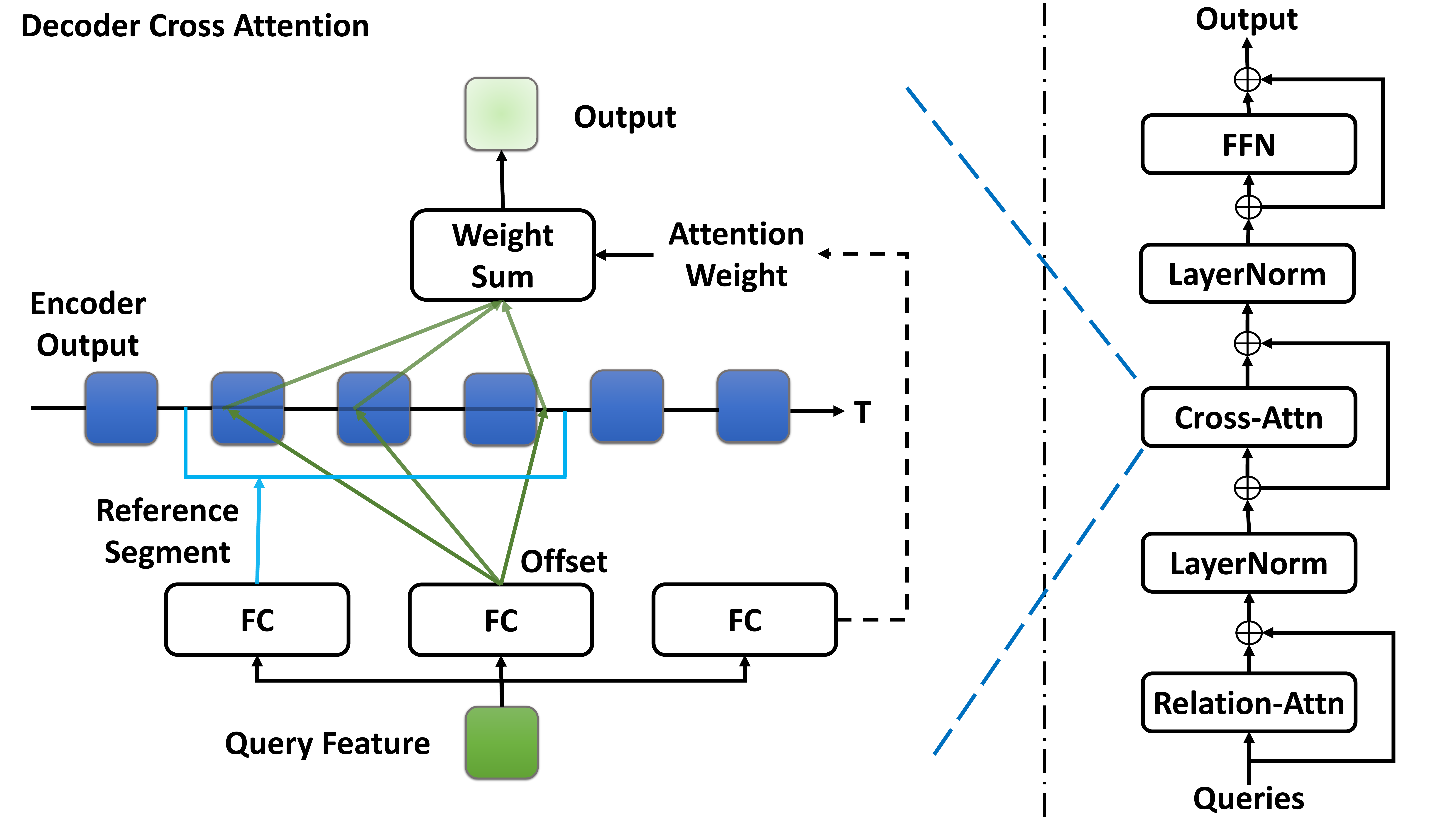}
  \caption{Illustration of the Deformable Cross Attention module.}
  \label{decoder}
  \vspace{-0.5cm}
\end{figure}

In the following, we elaborate on the deformable cross-attention module.
As \figref{decoder} showed, reference segment, offset position, and attention weights are predicted by three fully-connect layers, based on which the network samples sparse features to update the query feature at each decoder layer. There are two main differences in the deformable attention module between the encoder and decoder. 
First, the inputs and outputs are different. The input of the cross-attention in the decoder is the queries, while the input of the encoder is video features. The second difference is the reference segment. In the encoder, temporal offsets for each frame are sampled only around that frame. Whereas for the cross-attention module, an additional reference segment length is predicted for each query feature, and the offsets are normalizes such that the sampled frames are always in the segment. 
%The reference segment can also be preset to a fixed value, like the predicted segment of the last layer, and we do this for the layers after the first one for the segment refinement in practice. 
%Moreover, to compute the ACE-\textit{dec} loss, we also additionally feed the ground-truth segment as the reference segment to train classifier. 

\section{Architecture and Training Detail}
For THUMOS14, following~\cite{xu2020g}, we use the TSN network~\cite{wang2018temporal} pre-trained on Kinetics~\cite{kay2017kinetics} to extract features, which are then down-sampled every five frames. 
Each video feature is cropped in sequence with a window size 256, and two adjacent windows will have 192 overlapped features with a stride rate of 0.25. In the training phase, ground-truth cut by windows over  $75\%$ duration will be kept, and all empty windows without any ground-truth are removed. Finally, all ground-truth coordinates are re-normalized to the window coordinate system. 
 we set $L_q=40$, $L_E=2$, $L_D=4$ for the number of queries, encoder layer and decoder layer, respectively. Each deformable attention module will sample 4 temporal offsets for computing the attention. The hidden layer dimension of the feedforward network is set to 1024, and the other hidden feature dimension in the intermediate of the network is all set to 256. The pair-wise IoU threshold $\tau$ and feature similarity threshold $\gamma$ in \cls module are set to 0.5 and 0.2, respectively. 
%\subsubsection{ActivityNet-1.3.} 
For ActivityNet, the pre-trained TSN network by Xiong~\etal~\cite{xiong2016cuhk} is adopted to extract features. Then each video feature downsamples every 16 frames, and the resultant feature will be rescaled to 100 snippets using linear interpolation. 
We only do video-level detection instead of window-level. 
We set the $L_q=60$, $L_E=3$, $L_D=4$. We sample 4 temporal offsets for the deformable module. 
The dimension of hidden features is set to 256, and we set the pair-wise IoU threshold $\tau$ and feature similarity threshold $\gamma$ to 0.9 and -0.2, respectively. Following previous works~\cite{xu2020g,zeng2019graph,zhao2020bottom,yang2020revisiting}, we combined the Untrimmed-Net video-level classification results~\cite{wang2017untrimmednets} with our classification score.

\section{Visualization of the Classification Loss}
\begin{figure}[]
    \centering
    \includegraphics[width=0.4\linewidth]{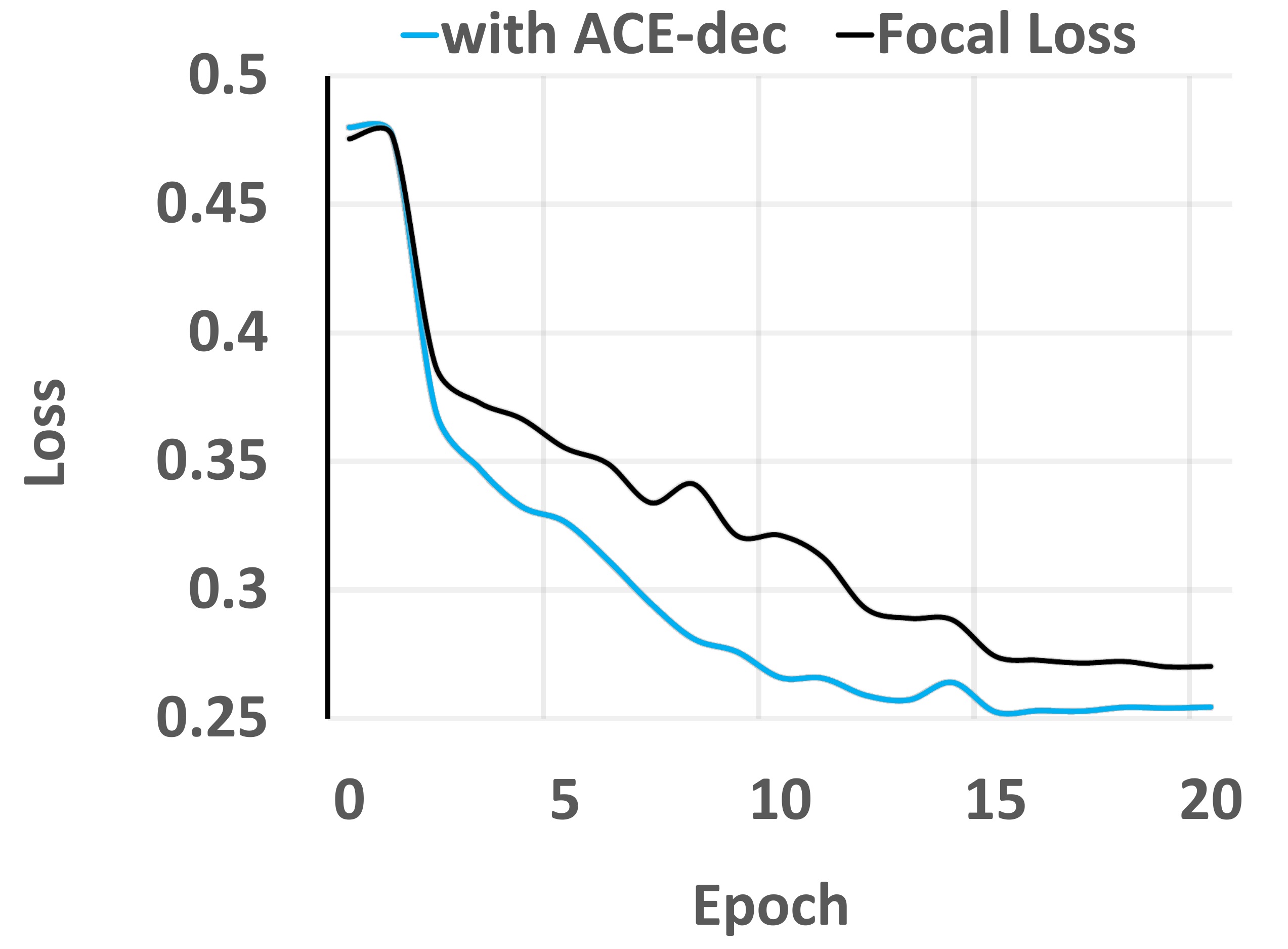}
  \caption{Visualization of the test classification loss. We record the testing loss with or without ACE-\textit{dec} loss during training}
  \label{loss}
%   \vspace{-0.2cm}
\end{figure}

To further demonstrate the effect of ACE-\textit{dec} loss, we compute the classification loss for the Activitynet-1.3 test set. As \figref{loss} shows, compared to the Focal Loss, the ACE-\textit{dec} loss improves not only the convergence speed but also the accuracy.

% \section{Length Analysis for Datasets}
%  \begin{figure}[H]
%     \centering
%     % \setlength{\abovecaptionskip}{-0.2cm}
%     \includegraphics[width=0.8\linewidth]{figure/statis.pdf}
%     \label{statis}
%   \caption{Action length statistics of THUMOS14 and ActivityNet-1.3.}
% %   \vspace{-0.1cm}
% \end{figure}

% %   \vspace{-0.1cm}
%  As \figref{statis} shows, we count the proportion of the action segments in different lengths, where the length is quantified into ten bins. We observe that the action instances in ActivityNet-1.3 have much more significant variance in action length than THUMOS14. However, for the DETR-based methods, input-independent queries are still difficult to model all the cases efficiently. We consider introducing video-related input for the decoder in future work. 
 
% ---- Bibliography ----
%
% BibTeX users should specify bibliography style 'splncs04'.
% References will then be sorted and formatted in the correct style.
%
\bibliographystyle{splncs04}
\bibliography{egbib}

\end{document}